\pdfoutput=1

\documentclass[11pt]{article}

\usepackage{ACL2023}

\usepackage{times}
\usepackage{latexsym}

\usepackage[T1]{fontenc}

\usepackage[utf8]{inputenc}

\usepackage{microtype}

\usepackage{inconsolata}

\usepackage{amsmath}
\usepackage{multirow}
\usepackage{makecell}
\usepackage{tabularx}
\usepackage{graphicx}
\usepackage{float}
\usepackage{tgtermes}
\usepackage[T1]{fontenc}
\usepackage{amsfonts}
\usepackage{comment}
\usepackage[symbol]{footmisc}
\renewcommand{\thefootnote}{\fnsymbol{footnote}}

\title{Critic-Guided Decoding for Controlled Text Generation}

\author{Minbeom Kim$^{1\ast}$ $  $ Hwanhee Lee$^{1}$ $  $ Kang Min Yoo$^{1,2}$ $  $ \textbf{Joonsuk Park$^{2,3}$ $  $ Hwaran Lee$^{2\dagger}$ $  $ Kyomin Jung$^{1\dagger}$} \\
    $^{1}$Seoul National University $  $
    $^{2}$NAVER AI Lab $  $
    $^{3}$University of Richmond\\
    \texttt{\{minbeomkim, wanted1007, kjung\}@snu.ac.kr}\\
    \texttt{\{kangmin.yoo, hwaran.lee\}@navercorp.com}, \texttt{park@joonsuk.org}\\
}

\begin{document}
\maketitle

\footnotetext{\textsuperscript{$\ast$} Work done during an internship at NAVER AI Lab.}
\footnotetext{\textsuperscript{$\dagger$} Corresponding authors.}

\renewcommand*{\thefootnote}{\arabic{footnote}}
\setcounter{footnote}{0}

\newcommand{\hwaran}[1]{\textcolor{violet}{#1}}

\begin{abstract}


Steering language generation towards objectives or away from undesired content has been a long-standing goal in utilizing language models (LM). Recent work has demonstrated reinforcement learning and weighted decoding as effective approaches to achieve a higher level of language control and quality with pros and cons. In this work, we propose a novel critic decoding method for controlled language generation (CriticControl) that combines the strengths of reinforcement learning and weighted decoding. Specifically, we adopt the actor-critic framework to train an LM-steering critic from non-differentiable reward models. And similar to weighted decoding, our method freezes the language model and manipulates the output token distribution using called critic, improving training efficiency and stability. Evaluation of our method on three controlled generation tasks, namely topic control, sentiment control, and detoxification, shows that our approach generates more coherent and well-controlled texts than previous methods. In addition, CriticControl demonstrates superior generalization ability in zero-shot settings. Human evaluation studies also corroborate our findings.

\end{abstract}

\section{Introduction}


Recent breakthroughs in large language models (LMs) have allowed them to generate sentences that are as much more natural and plausible as real-world text ~\cite{radford2019language, brown2020language, kim2021changes}.
However, there are potential risks in simply generating text by following the training data distribution because the real-world data include harmful, offensive, or socially biased expressions \cite{lu2022quark, liu2021dexperts, gehman2020realtoxicityprompts, hosseini2017deceiving}.
The LMs also may generate misinformation, the so-called Hallucination problem, leading to untrustworthy generation results \cite{holtzman2018learning}.
To mitigate these risks and accomplish the text generation goals, many methods for controlling the language models have been proposed \cite{keskar2019ctrl, krause2020gedi, yang-klein-2021-fudge, dathathri2019plug}, but effective and efficient control ways remain a crucial challenge.


\begin{figure}[t!]
\centering
\includegraphics[width=0.9\columnwidth]{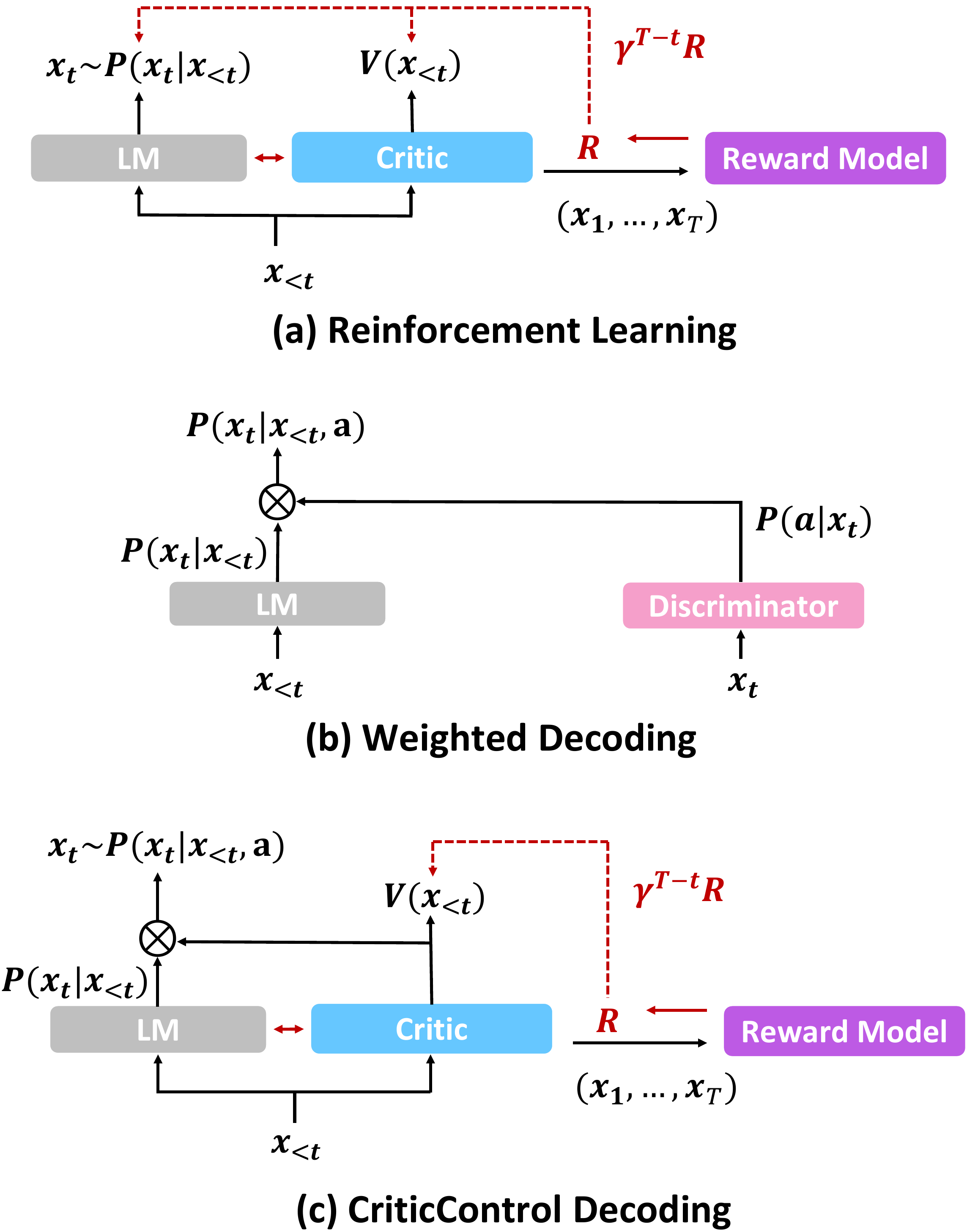}
\caption{Comparison of three controlled text generation methodologies. (a) and (c) have commons in sequential decision making. Both (b) and (c) have freezed LM and the `plug-and-play' flexible controllability.}
\label{fig:intro}
\vspace{-5mm}
\end{figure}
Previous research on controlled text generation can be categorized into two major approaches: reinforcement learning (RL) and weighted decoding (WD). 
First, in the RL framework, generating a word in a sequence at each time is regarded as a sequential decision-making process according to the LM’s action policy $\pi$. Specifically, in the widely used Actor-Critic framework, a pre-trained LM is regarded as the \textit{actor} for the sequential decision-making $p_\pi(x_t|x_{<t})$ and trained to maximize the expected rewards such as ROUGE~\cite{paulus2017deep}, BLEU~\cite{wu2016google} and even human reference feedback~\cite{stiennon2020learning, wu2021recursively}. And an additional value-predictor network \textit{critic} is trained to evaluate the current states (i.e., generated text up-to-the-current) by these rewards. However, training the actor-networks (i.e., LMs) often suffers from the noise gradient estimation~\cite{greensmith2004variance}, leading to unstable training and mode collapse~\cite{upadhyay2022efficient}.
Besides, optimizing the entire large LM actor network for each attribute results in computational costs and memory inefficiency for real-world application scenarios~\cite{guo2021text}.

On the other hand, weighted decoding~\cite{holtzman2018learning, dathathri2019plug, kumar2021controlled} takes a more applicable approach to address these limitations. Instead of fine-tuning the full language model for the target attribute, these approaches use external discriminators to adjust only the final output probability distribution, leaving the underlying language model freezed~\cite{liu2021dexperts} as shown in Figure~\ref{fig:intro} (b). Specifically, to control the attribute with the pure output $p(x)$ from language models, they compute the approximation of bayesian decomposition $p(x|a) \propto p(x)p(a|x)$ with $p(a|x)$ from external discriminators~\cite{yang-klein-2021-fudge, krause2020gedi}. 
These methods use an external discriminator for each attribute, and this plug-and-play structure is more memory-efficient and also makes a more simple training procedure~\cite{gu2022distributional} than RL approaches.
However, since they are independently trained with the freezed language model, these discriminators often negatively impact the quality of the generated texts such as fluency and diversity~\cite{lu2022quark}.



In this paper, we propose a novel controlled text generation algorithm called Critic-Guided Decoding for Controlled Text Generation (CriticControl) that re-weight the word distribution $p_\pi(x)$ from LM with predicted state-values from the critic network as depicted in Figure~\ref{fig:intro}.
Particularly, for each word $x_t$ at time-step $t$, we calculate the ratio of the next state-value with the word to the current state-value (i.e., $\alpha(x_t,x_{<t}) = \frac{V_\pi(x_{\le t})}{V_\pi(x_{<t})}$). Then, the CriticControl decodes language under the reward-pursuing word distribution, which is defined as $p_{\pi}(x_{t}|x_{<t}) = \alpha(x_{t},x_{<t}) p_{\pi}(x_t|x_{<t}) $.
As a result, the CriticControl raises the likelihood of words that increases the value of the next state over the current state, while it lowers the others.
Since the critic network is trained with the frozen backbone of LM, the proposed method is able to accomplish training efficiency and stability. Moreover, the different critic networks can be utilized in a plug-and-play manner depending on targeting attributes or reward models.

In the experiment, the efficacy of the CriticControl algorithm is demonstrated on three controlled text generation tasks: topic control, sentiment control, and toxicity control tasks. For all tasks, we show that CriticControl consistently outperforms previous methods in terms of control success, fluency, and diversity. We also observe that the CriticControl is superior even for zero-shot control settings, implying that the methods can handle unseen control codes.
Finally, we show that the decoding module of the CriticControl can combine with a widely used \textit{top-k}, \textit{top-p} sampling methods ~\cite{holtzman2019curious, fan2018hierarchical} to produce more fluent and human-like text.

\section{Related Works}

\label{sec:relatedwork}

\textbf{Reinforcement Learning} RL and the adversarial training formulation was first proposed in the context of language generation as an auxiliary algorithm to mitigate exposure bias in the teacher-forcing training of sequences~\cite{ranzato2015sequence, wu2016google, hu2017toward}. The main motivation is to incorporate readily-available sequence-level reward signals into training, such as BLEU or ROUGE~\cite{paulus2017deep}.
The success of utilizing RL have been observed in a wide range of tasks, including summarization~\cite{paulus2017deep, wu2018learning, stiennon2020learning, ziegler2019fine}, dialog modeling~\cite{li2016deep, yi2019towards, jang2021gpt, upadhyay2022efficient}, neural machine translation~\cite{wu2018learning, nguyen2017reinforcement}, and style transfer~\cite{gong2019reinforcement, ziegler2019fine}. Furthermore, robustness of RL has allowed models to capture high-level human feedback~\cite{paulus2017deep, stiennon2020learning, sharma2021towards, snell2022offline}, which is out of current work’s scope.
Various RL approaches have been explored so far, such as REINFORCE~\cite{sutton1999policy, ranzato2015sequence, wu2016google, sharma2021towards, upadhyay2022efficient}, the actor-critic framework~\cite{bahdanau2016actor, nguyen2017reinforcement, jang2021gpt}, and PPO~\cite{schulman2017proximal, nakano2021webgpt, snell2022offline}. CriticControl is similar to the work of \cite{jang2021gpt}, generating critic-guided texts for enriching task-oriented dialogue datasets. Our work is the first to incorporate weighted decoding for sequence-level reinforcement learning.

\paragraph{Weighted Decoding}
Freezing the language model and controlling the output probability distribution to suit the purpose is being actively researched~\cite{ghazvininejad2017hafez, keskar2019ctrl, sudhakar-etal-2019-transforming}. This mainstream is useful for various purposes, such as forcing the model to generate text that conforms to certain stylistic or content-based constraints or mitigating aggressive and toxic expressions~\cite{kumar2021controlled, gu2022distributional, mireshghallah2022mix}. Previous works~\cite{holtzman2018learning, arora2022director} stabilize repetitive and self-contradictory language models with a cooperative discriminator. The plug-and-play LM(PPLM) \cite{dathathri2019plug} generates texts by plugging a steering layer $p(a|x)$ into the top of base language models $p(x)$. Then, the gradient from $p(a|x)$ updates iteratively the last hidden representation to desired attributes. It only needs a few layers per each attribute but requires a lot of iterative computations; FUDGE~\cite{yang-klein-2021-fudge} economizes $p(x|a)$ by bayesian decomposition $p(x|a) \propto p(x)p(a|x)$ with $p(a|x)$ computing classifier instead of gradient update methods. For better linguistic quality, GeDi~\cite{krause2020gedi} and DExperts~\cite{liu2021dexperts} take generative discriminator approaches with two LMs conflicting on the desired attribute. They interpret $p(a|x)$ as the ratio of disagreement between two models. In contrast, CriticControl is advantageous in taking a generative manner with only a steering layer, like PPLM.

\section{CriticControl}

\begin{figure*}[ht!]
\centering
\includegraphics[width=\textwidth]{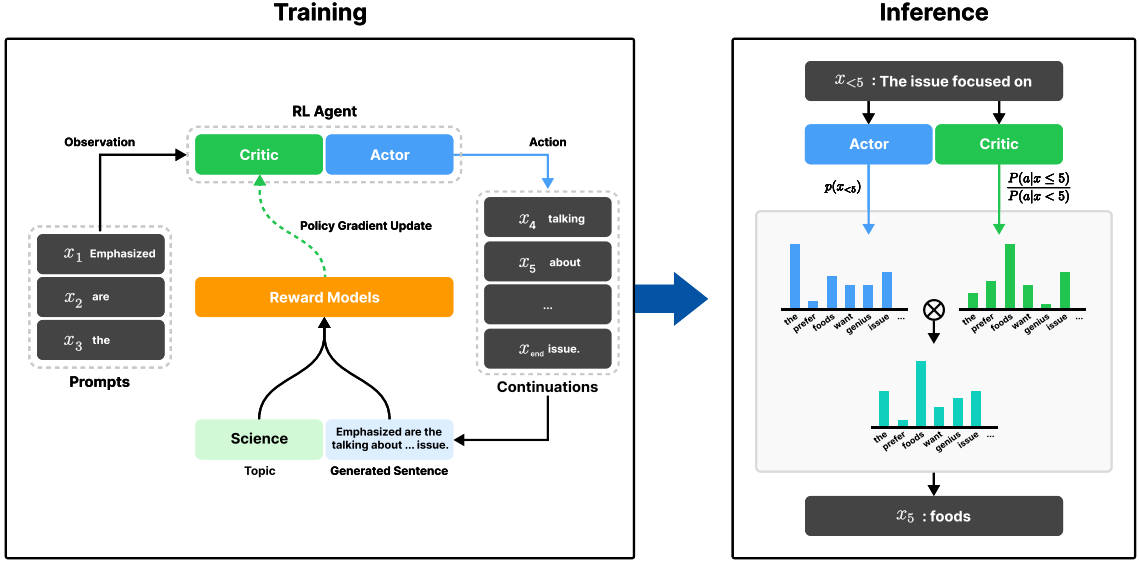}
\caption{The following example of a topic control task illustrates the overall flow of CriticControl. Language model completes the sentence in response to the prompt 'Emphasized are the', and the output is evaluated by the reward model, judging the relevance of the generated text to the specified topic. During the inference, critic modifies the output distribution of the language model to ensure that the generated text is appropriately related to the topic when choosing the next token.}
\label{fig:overview}
\vspace{-5mm}
\end{figure*}




CriticControl is a controlled text generation framework that consists of a freezed pre-trained language model (\textit{Actor network}) and an extra network for the state-value prediction (\textit{Critic network}). Given a part of the sentence, the actor tries to complete the attribute-aware continuations with the critic's support.

As a RL formulation, at each step $t$, states are given as a set of tokens $x_{1:t-1} = \{x_1, ..., x_{t-1}\}$ and the attribute $a$. The policy $\pi_\theta$ of the actor samples $x_t \in \mathcal{V}$ from the next tokens probability of $\pi_\theta(x_t|x_{t-1})$ as
\begin{equation}
\begin{aligned}
    P_\pi(x_t|x_{1:t-1}) = \mathrm{softmax}\left(\frac{y_t}{T}\right
),
\label{eq:softmax}
\end{aligned}
\end{equation}
output logits $y_t$ for words in vocabulary set $\mathcal{V}$ with temperature $T$ to experience more diverse trajectories. 
To train the policy network (i.e., LM), widely used optimization methods~\cite{sutton1999policy} take policy loss as $\nabla_\theta J(\theta) = \mathbb{E}_\pi [\sum_{t=1}^{end} A_t \nabla_\theta \ln \pi_\theta(x_t | x_{1:t-1})]$, where $A_t$ is the advantages function that measures how much the choice $x_t$ is better than the critic predicted, and the critic learns to make $A_t$ zero.
Note that, in our algorithm, we however freeze the actor model and only train the critic for flexible control as described in the following section.

\subsection{CriticControl Training}
Different from the previous supervised training regime, we design a simple text game for reinforcement critic learning. As shown in~\ref{fig:overview}, when the actor completes a given prompt, we let the reward model evaluate how well the completed sentence correlates with the attribute $a$. Then we give critic a reward $r_{end} = P_\pi(a|x_{1:end})$ to calculate the temporal difference (TD) error $\delta_{t} = r_{t} + \gamma V_\pi(x_{1:t+1}) - V_\pi(x_{1:t})$, consisting of $A_t$. 
This TD error generalizes $V_\pi(x_{1:end}) = P_\pi(a|x_{1:end})$ to $V_\pi(x_{1:t}) = P_\pi(a|x_{1:t})$~\cite{sutton2018reinforcement}. For unbiased and stable critic training, we take the generalized advantage estimation loss~\cite{schulman2015high} using the evaluation result $r_{end}$ and the text generation history $x_{1:end}$ as follows:
\begin{equation}
\begin{aligned}
     \mathcal{L}_{critic} = \sum_{t=1}^{end}(\sum_{i=0}^{end-t}(\gamma \lambda)^i \delta_{t+i})^2,
\label{eq:critic_loss}
\end{aligned}
\end{equation}
aligning critic’s prediction $V_\pi(x_t)$ to unbiased empirical future returns with reward discount factor $\gamma$, the re-weighted averaging factor $\lambda$. During the exploration in training, we adopt a highly stochastic actor strategy. Actor samples outputs of language models with high temperatures $T > 1$ as in Equation~\ref{eq:softmax} by mitigating hardly confident output distributions of logits $y_t$ over a vocab of tokens $\mathcal{V}$. And this leads to diverse text generation, and critic can experience more diverse samples in each episode. By repeating this simulation, we expect the critic to learn which decisions of language models will lead to a promising future.

\subsection{Text Generation with CriticControl}

Generating human-like texts commonly requires stochastic decoding strategies. They truncate the unreliable long-tail on the probability distribution for sampling only on realistic token candidates~\cite{holtzman2019curious, fan2018hierarchical}. However, adjusting all probabilities in the vocabulary is computationally very inefficient. Therefore, to achieve stochasticity and computational efficiency simultaneously, CriticControl steers only a small subset $\mathcal{V'} \subset \mathcal{V}$ with exact distribution shift 
\begin{equation}
\begin{aligned}
    P_{\pi}(x_t \vert x_{<t}, a) &= \frac{P_{\pi}(x_t, x_{<t}, a)}{P_{\pi}(x_{<t}, a)}\\
    &= \frac{P_\pi(a|x_{\le t})}{P_\pi(a|x_{<t})} P_\pi(x_t|x_{<t}).
\label{eq:stepwise_shift}
\end{aligned}
\end{equation}
to align the probability scale of adjusted words $x \in \mathcal{V'}$ and non-adjusted words $x \notin \mathcal{V'}$. Through this computation, CriticControl combines beam search and various sampling methods. In experiments, CriticControl adopts both top-k sampling and nucleus sampling by adjusting the top 10 word probabilities of $\mathcal{V'} \subset \mathcal{V}$.

\section{Experiments}

To evaluate the effectiveness of CriticControl, we conduct experiments on a variety of controlled text generation tasks, including topic control, sentiment control, and detoxification. 

\subsection{Topic Control}
\label{sec:topic_control}

\begin{table*}[th!]
\centering
\begin{tabular}{lcccccc}
\Xhline{3\arrayrulewidth}
&  & \multicolumn{2}{c}{\textbf{Fluency}} & \multicolumn{3}{c}{\textbf{Diversity}} \\
\multirow{-2}{*}{\textbf{Model}} & \multirow{-2}{*}{\textbf{Success}}      & Perplexity                 & Grammar                & Dist-1           & Dist-2           & Dist-3          \\
\hline
GPT2 - medium                                     & 0.16              & 14.06                      & 0.74                   & 0.29             & 0.70             & 0.88            \\
WDEC                                            & 0.49              & 67.53                      & 0.59                   & 0.16             & 0.42             & 0.85            \\
PPLM                                            & 0.45              & 62.66                      & 0.78                   & 0.35             & \textbf{0.78}             & \textbf{0.92}            \\
FUDGE                                           & 0.78              & 69.08                      & 0.79                   & 0.34             & 0.75             & 0.91            \\
CriticControl                                     & \textbf{0.89}              & \textbf{17.19}                      & \textbf{0.83}                   & \textbf{0.49}             & 0.76             & 0.90            \\
\hline
CriticControl - small                              & 0.85              & 16.88                      & 0.83                   & 0.47             & 0.73             & 0.89            \\
CriticControl - large                              & 0.92              & 17.58                      & 0.84                   & 0.51             & 0.77             & 0.91            \\
CriticControl - XL                                 & \textbf{0.94}              & 17.69                      & 0.83                   & 0.51             & 0.77             & 0.91 \\
\hline
CriticControl - Zero shot  & \textbf{0.73}              & 17.55                      & 0.85                   & 0.49             & 0.76             & 0.90 \\
\hline
\end{tabular}
\caption{Automatic Evaluation Results of Topic control language generation experiments from GPT2. The first section shows comparison over baselines steering freezed GPT2-medium. Baseline results are adopted from FUDGE with \textbf{bold} key results. In the second, CriticControl-[size] indicates the size of the freezed GPT2. We conducted an ablation study on CriticControl's efficacy according to size of GPT2. The last is an experiment on the generalization ability of CriticControl. Other than 7 training topics, we verify how robust control is possible even on completely new topics.}
\label{topic_auto}
\end{table*}

We conduct experiments on topic control tasks to verify the efficacy of the Critic-Guided method in generating text that is controllable, fluent, and able to generalize to new topics without training. These experiments aim to generate topic-related text starting from a prompt that is totally irrelevant to the topic. To provide a comprehensive comparison with previous work, we evaluate the performance of our approach and the baselines using both automatic evaluation metrics and human evaluation for the quality and diversity of the generated text.

\subsubsection{Implementation Details}

To control a language model for universal topics, we adopted \textit{BART-Large-MNLI}~\cite{lewis2019bart}, language models for measuring universal relevance, as a general reward model rather than a binary classifier. When GPT2 generates 80 continuations from the prompt, the reward model evaluates how relevant the generation is to the desired topic, and the critic learns to predict this relevance-reward. We follow the training setup of PPLM~\cite{dathathri2019plug}, training value predictor with seven topics (computers, space, military, legal, politics, science, and religion)). We take temperature $T=2$ during sequential samplings, exploring various text trajectories. For inference, CriticControl gets repetition-penalty with greedy decoding strategy. We follow previously reported baselines of \cite{yang-klein-2021-fudge}, including pure \textit{GPT2-medium} for naive generated texts, PPLM~\cite{dathathri2019plug}, WDEC and FUDGE~\cite{yang-klein-2021-fudge}. Additionally, we conduct an ablation study on language model scales from \textit{GPT2-small} to \textit{GPT2-XL}, and generate texts on zero-shot topics(Foods, Games, Donald Trump, ...) for discussing generalization capability.

\begin{table}[ht!]
\centering 
\begin{tabular}{p{0.9\linewidth}}
\hline
\textbf{Music:} Emphasised are the words "instrument" and "instrumentals" in the title.  The song is a cover of the song "I'm a Man" by the band The Beatles.  "I'm a man" is a reference to the song "Man of the World" by the British band The Beatles, which was written by John Lennon and Paul McCartney. \\
\hline
\textbf{Foods:} The issue focused on the use of the term "organic" in the food industry.  This issue focused on a new USDA regulation that requires food companies to label their products as "organic" if they meet certain criteria. The regulation was passed in 2010, but the Food and Drug Administration (FDA) has yet to issue a final rule. \\
\hline
\end{tabular}
\caption{The zero-shot control output of topic-prompt pairs {Music, Emphasised are} and {Foods, The issue focused on}. 'Beatles' and 'USDA regulation' are hard to observe during training; however, CriticControl samples those words naturally if given topics are related.}
\label{intro_zero_shot_example}
\vspace{-5mm}
\end{table}






\begin{figure*}[ht!]
\centering
\includegraphics[trim=250 450 50 420, clip,width=\linewidth]{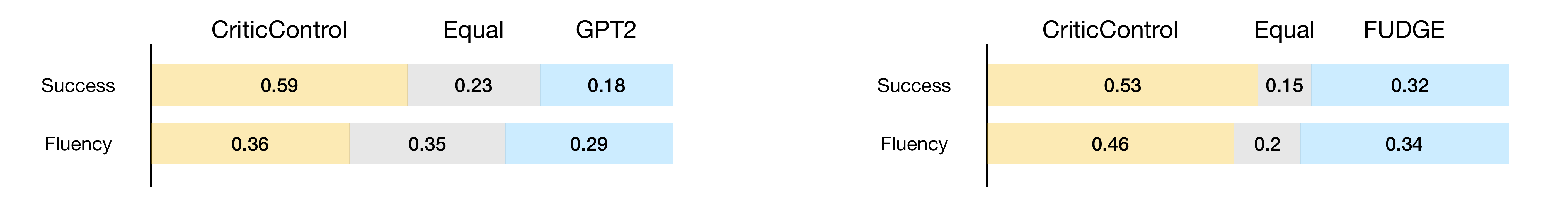}
\caption{Human evaluation results for topic control. This experiment compares human preferences between two generations under the same prompts and topics, CriticControl vs. GPT2-medium and CriticControl vs. FUDGE.}
\label{fig:topic_preference}
\end{figure*}

\subsubsection{Automatic Evaluation} 
We evaluate the quality of the generated text in terms of controllability, fluency, and diversity. For controllability, it is difficult to directly compare the quality of our approach with that of previous baselines, as they reported the controllability based on their own "bag of words" and directly optimize for specific predefined topic-related words. In contrast, our approach performs more general optimization and does not pre-define topic-related words, which leads to a significant discrepancy in automatic evaluation metrics on controllability. Hence, we decided to rely solely on human evaluation to assess the success of our topic control experiments. We describe the detailed design of this evaluation method in the human evaluation section. And we measure the fluency using two metrics: perplexity, which is calculated using the \textit{GPT2-XL} language model, and grammaticality, which is determined using the \textit{Roberta-based CoLA} model~\cite{warstadt2019neural}~\footnote{https://huggingface.co/textattack/roberta-base-CoLA}. Finally, we measure diversity~\cite{li2015diversity} using distinct n-grams normalized by text length, reporting distinct unigrams, bigrams, and trigrams as Dist-1, Dist-2, and Dist-3 scores, respectively.

\paragraph{Results} As shown in Table~\ref{topic_auto}, CriticControl exhibits high text quality that significantly outperforms other baselines on a variety of on-domain topics. In particular, CriticControl is able to well preserve the linguistic characteristics of the frozen GPT2 by training the critic using a sequential decision-making approach, in contrast to the baselines that rely on external classification models and suffer from poor linguistic fluency and diversity, as verified by the perplexity and the grammar scores. Additionally, the unigram score demonstrates the effectiveness of CriticControl, which generates texts without any pre-defined "bag of words" about topics and this makes the system freely choose the topic-related words, unlike WDEC, PPLM, and FUDGE.

\subsubsection{Human Evaluation} 

Different from binary tasks such as sentiment control and detoxification, it is hard to construct a universal metric of `being on-topic.' In addition, there is a limitation to solely relying on automated evaluation for measuring text quality. Therefore, we run human evaluations through Amazon Mechanical Turk to validate our method on three experiments; topic control success rate of CriticControl, preference test of `CriticControl versus GPT2 and FUDGE', and topic control success on the unseen topic (i.e. zero-shot setting). 
For the preference test, we hire three annotators for each comparison pair and ask two questions \textit{1) Which sentence is more related to the given topic?} and \textit{2) Which sentence is more fluent?}

\begin{figure}[h!]
\centering
\includegraphics[trim=0 0 100 0, clip,width=\columnwidth]{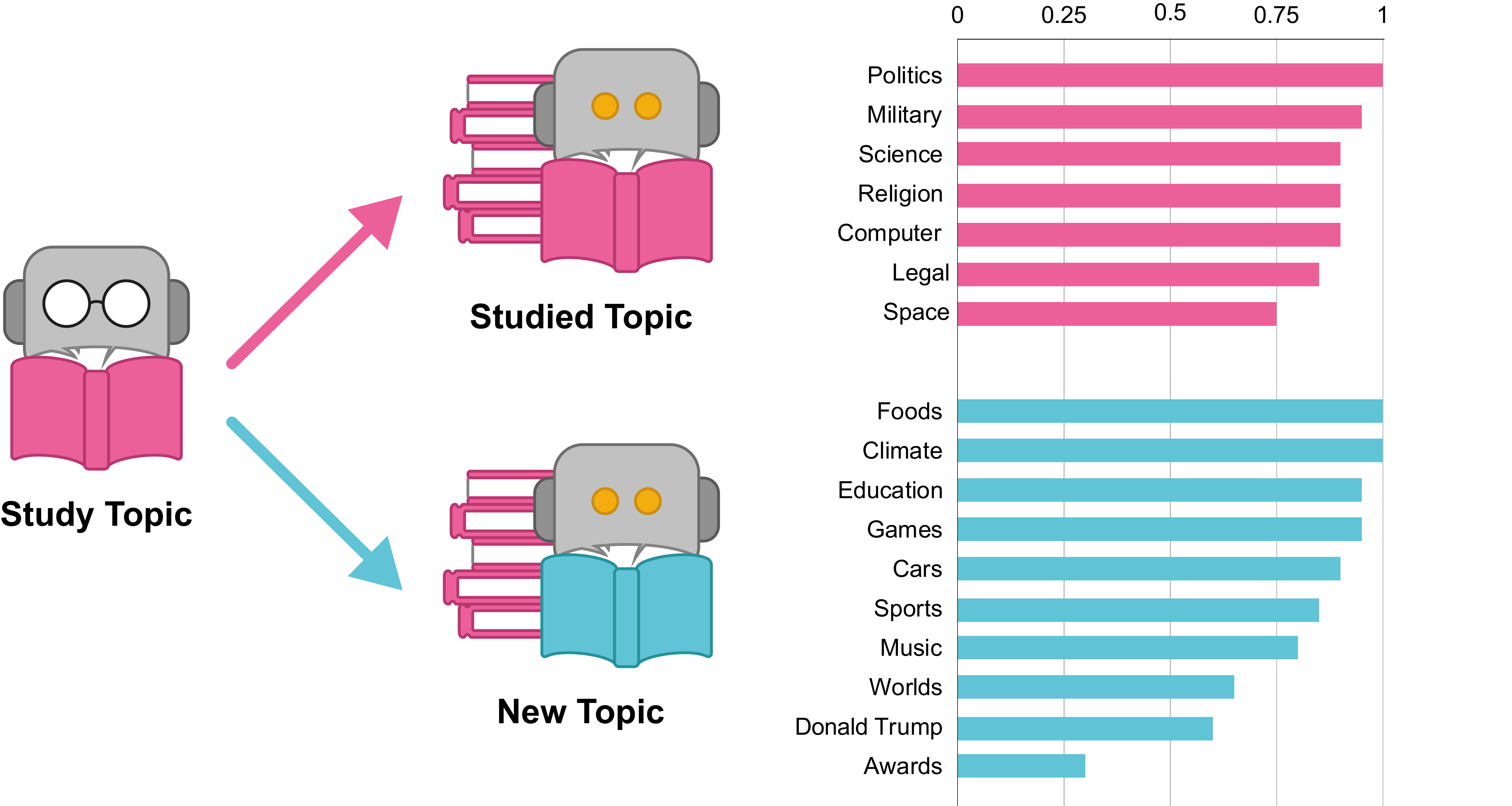}
\caption{The CriticControl success rate per topic. It is examined for both the topic used for training and a new, previously unseen topic. The results indicate consistent performance on zero-shot topics.}
\label{fig:zero_shot_score_per_topic}
\end{figure}

\paragraph{Results} According to results in~\ref{topic_auto}, CriticControl achieves a superior control success rate, and this tendency is proportional to the size of the GPT2 model. And even small-sized CriticControl beats the FUDGE in terms of topic controllability. Furthermore, in the preference test, CriticControl also outperforms baselines on both success and fluency as shown in Figure~\ref{fig:topic_preference}. Also, one of the most promising results is that GriticGuide almost completely control unseen topics other than seven training domains in the zero-shot topic experiments. While preserving GPT2's text quality, CriticControl achieves a success rate almost as high as the topic used in in-domain, boasting great generalization ability as shown in \ref{fig:zero_shot_score_per_topic} and \ref{intro_zero_shot_example}. These experimental results verify that this reward-driven controllable text generation system is suitable for sequential decision-making and obtains promising generalization ability by taking universal reward models.

\subsection{Sentiment Control}

\begin{table*}[th!]
\centering
\begin{tabular}{lcccccc}
\Xhline{3\arrayrulewidth}
& \textbf{Success} & \multicolumn{2}{c}{\textbf{Fluency}} & \multicolumn{3}{c}{\textbf{Diversity}} \\
\multirow{-2}{*}{\textbf{Model}} & Positiveness      & Perplexity                 & Grammar                & Dist-1           & Dist-2           & Dist-3          \\
\hline
GPT2 - medium  & 0.57   & 11.91     & 0.78      & 0.25   & 0.63  & 0.78            \\
PPLM   & 0.60   & 142.11     & 0.73    & 0.22  & 0.61  & 0.72            \\
CC-LM    & 0.76  & 15.79   & 0.72   & 0.28   & 0.70  &0.82            \\
GeDi  & 0.84     & 38.94     & 0.76      & 0.27     & 0.77     & 0.89            \\
CriticControl   & \textbf{0.90}    & \textbf{12.97}   & \textbf{0.87}                   & \textbf{0.31}             & \textbf{0.84}             & \textbf{0.92}            \\
\hline
PPO   & 0.94 & 13.43    & 0.84    & 0.32   & 0.86 & 0.93            \\
PPO - CriticControl     & \textbf{0.99}    & 13.44     & 0.80    & 0.32  & 0.85  & 0.93            \\
\hline
\end{tabular}
\caption{Automatic Evaluation Results of Sentiment control language generation experiments from GPT2. The first is about comparison over steering freezed GPT2-medium with each guided decoding; PPLM, CC-LM, GeDi, and CriticControl. The last is for verifying the control potential to steer unfreezed language models. This experiment compares GPT2-medium finetuned on PPO and those PPO with CriticControl. The key results are in \textbf{bold}.}
\label{sentiment_auto}
\end{table*}

\begin{figure*}[ht!]
\centering
\includegraphics[trim=250 450 50 420, clip,width=\linewidth]{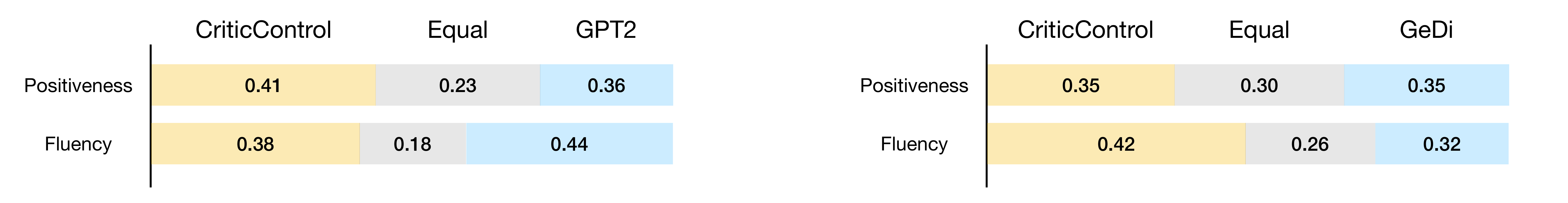}
\caption{Human evaluation results for sentiment control. This experiment draws comparisons on human preferences between two continuations after the same prompts, CriticControl vs. GPT2 and CriticControl vs. GeDi.}
\label{fig:sentiment_preference}
\end{figure*}

Next, we explore CrilticControl's ability to steer text generation toward a specific sentiment. This sentiment control task aims to steer the model to complete positive movie reviews with any emotional prompt. We adopt the IMDB movie review dataset, containing highly polar (positive or negative) reviews of 2.5K for training and 2.5K for testing. We modified the sentences to prompts, truncating them to 8 starting tokens. We generate 25 continuations for each prompt for automatic and human evaluation for all of the baseline systems.

\subsubsection{Implementation Details}

Since this task is to manipulate the emotions of movie reviews, all baselines compare the controlled text from \textit{GPT2-medium} finetuned on IMDB movie review dataset. During training procedure, LM completes texts from GPT2 without any shifts, starting with positive and negative IMDB prompts. Then, the critic observes these experiences and learns rewards generated from the reward model, \textit{distilBERT}~\cite{sanh2019distilbert} sentiment classifier finetuned on 2.5k reviews in IMDB dataset. Finally, CriticControl generates critic-guided text using nucleus sampling with a probability of 0.9 on test datasets. All baselines generate and compare their own `guided' texts from the 2.5k test IMDB prompts. Our discussion starts from GPT2 - medium, the basic baseline for sentiment control. For discussing PPLM, we retrain IMDB sentiment classifier for gradient updates. Then, PPLM decodes greedily on updated latent representation $H_t + \Delta H_t$. GeDi consists of \textit{GPT2-XL} and two polar generative discriminators, CC-LMs~\cite{krause2020gedi} finetuned on `positive' or `negative' IMDB movie reviews. For a fair comparison, we downgrade GeDi's language model to \textit{GPT2-medium} finetuned on IMDB. We discuss both `Positive' CC-LM and GeDi for the sentiment control experiment. We add an experiment to answer the question, "Could reinforcement-learned models be critic-guided to achieve goals more appropriately?" To verify this potential, we finetune PPO \cite{schulman2017proximal} on unfreezed \textit{GPT2-medium} by our reward model, and compare naive PPO and PPO with CriticControl.





\subsubsection{Automatic Evaluation}

We evaluate total $2.5k \times 7 = 17.5k$ generations about positiveness, fluency, and diversity. We use the positiveness metric as \textit{distilBERT} classifier finetuned on IMDB dataset, and report the percentage of generations classified to `positive' using this model. Same as topic control task, we measure fluency in two ways, the perplexity of \textit{GPT2-XL} and grammaticality of CoLA grammaticality model. We also compute Diversity following the previous sections~\ref{sec:topic_control}.

\paragraph{Results} As shown in Table~\ref{sentiment_auto}, CriticControl significantly outperforms the other baselines on the success rate of sentiment control, fluency, and diversity metrics. Also, we observe that generative controllers such as CriticControl, GeDi, and CC-LM demonstrate better performance than PPLM. Among them, GeDi improves control performance by generating guided texts through the use of contradicting positive and negative CC-LMs. However, the differing text generation strategies of GeDi and CC-LMs leads to a reduction in text quality, as indicated by the perplexity score. On the other hand, CriticControl trains the critic in a generative manner and allows the actor and critic to share the same experience, resulting in the best performance on all metrics. And both the topic and sentiment control experiments show the effectiveness of CriticControl in improving grammatical correctness compared to naive GPT2. We explain that this is because CriticControl increases the amount of information within the region identified by the reward model, whereas naive GPT2 does not. Furthermore, our additional experiment on PPO in Table~\ref{sentiment_auto} shows that CriticControl can improve the performance of RL-finetuned language models, not just freezing language models. 
Overall, our results show that CriticControl is promising for extending the use of reinforcement learning in downstream tasks.

\subsubsection{Human Evaluation} 

For human evaluation, we conduct preference tests by comparing CriticControl with GPT2-medium and GeDi. We randomly select 200 samples from the test set and ask annotators to indicate 1) which sentences are more positive and 2) which are more linguistically fluent. As in the topic control experiment, we take the majority vote of 3 annotators for each comparison.

\begin{table*}[ht!]
\centering
\begin{tabular}{lcccccc}
\Xhline{3\arrayrulewidth}
& \textbf{Success} & \multicolumn{2}{c}{\textbf{Fluency}} & \multicolumn{3}{c}{\textbf{Diversity}} \\
\multirow{-2}{*}{\textbf{Model}} & Toxic prob & Perplexity  & Grammar                  & Dist-1           & Dist-2           & Dist-3          \\
\hline
GPT2 - large & 0.520   & 11.31  & 0.84  & 0.58   & 0.85  & 0.85            \\
PPLM  & 0.518   & 32.58 & 0.75 & 0.58 & \textbf{0.86}  & 0.86            \\
DAPT  & 0.360  & 31.21 & 0.71 & 0.57 & 0.84  &0.84            \\
GeDi & 0.217     & 60.03 & 0.79 & \textbf{0.62}     & 0.84     & 0.83            \\
DExperts &  0.128     & 32.41 & 0.76 & 0.58     & 0.84     & 0.84            \\
CriticControl &  \textbf{0.081}    & \textbf{17.02} & \textbf{0.81} & 0.56             & 0.84             & \textbf{0.87}            \\
\hline
\end{tabular}
\caption{This experiment presents the results of automatic evaluations for detoxification. The results are compared with baselines for steering a freezed GPT2-Large model. Baseline results are adopted from the FUDGE study, with key results highlighted in \textbf{bold}. The objective of this experiment is to verify the effectiveness for reducing the toxicity of generated text, using a variety of automatic evaluation metrics.}
\label{toxic_auto}
\end{table*}

\begin{figure*}[ht!]
\centering
\includegraphics[trim=250 450 50 420, clip,width=\linewidth]{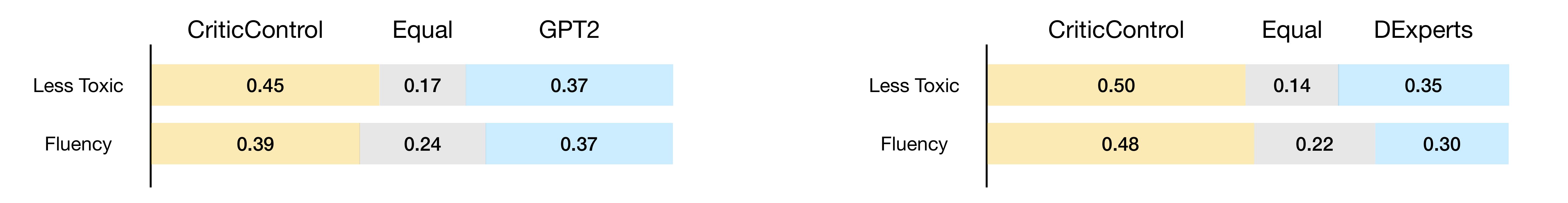}
\caption{Human evaluation results for detoxification. This experiment shows comparisons on preferences between two continuations after the same prompts, CriticControl vs. GPT2 and CriticControl vs. DExperts.}
\label{fig:toxic_preference}
\vspace{-3mm}
\end{figure*}

\paragraph{Results} Our initial experimental results with GPT2, as shown in Figure \ref{fig:sentiment_preference}, indicate that CriticControl is successful at generating positive text. However, because CriticControl is forcing negative prompts to be positive, the fluency of the generated text is poorer than that of a naive GPT. On the other hand, when we compare CriticControl to GeDi, we see that CriticControl has much better fluency while still maintaining a high level of positivity. 
However, we observe that people show the same preference for positivity among the two models, and choose equality for considerable samples, different from the automation evaluation results. We explain that this is because people have difficulty choosing more positive sentence among two positive sentences, unlike distinguishing between positive and negative sentences.

\subsection{Detoxification}

Large Language models can generate offensive or biased responses when given seemingly neutral prompts. One way to address this issue is using CriticControl to reduce the likelihood of toxic text at each step. By doing so, CriticControl can effectively decrease the amount of toxic text produced without negatively impacting the overall quality of the generated text, as measured by both human and automatic evaluation.

\subsubsection{Implementation Details}

In order to fairly compare our approach with other methods for controlled text generation, we use \textit{GPT2-Large} as our base language model. We train the reward model using \textit{BERT-based classification}~\cite{devlin2018bert} models on a dataset from the Jigsaw Unintended Bias in Toxicity Classification Kaggle challenge 
\footnote{\url{https://www.kaggle.com/c/jigsaw-unintended-bias-in-toxicity-classification}}. All evaluated generations start from the RealToxiciyPrompts dataset~\cite{gehman2020realtoxicityprompts}, which consists of $100K$ prompts designed to elicit toxic responses. And we follow the experimental setup used by DExperts. During critic training, we use $90K$ toxic and non-toxic prompts from the train set with our reward model. For evaluation, we use the same $10K$ non-toxic test prompts as DExperts and generate text using top-10 sampling. We also include reported baselines from DExperts~\cite{liu2021dexperts}, including naive \textit{GPT2-Large}, PPLM, DAPT~\cite{gururangan2020don}, GeDi, and DExperts.

\subsubsection{Automatic Evaluation} 

In this experiment, we evaluate a total of $10k$ generated sentences for toxicity, fluency, and diversity. We use the Perspective API~\cite{hosseini2017deceiving} \footnote{\url{https://perspectiveapi.com/}} to classify sentences based on the most toxic token. As in the previous section on topic control \ref{sec:topic_control}, we measure fluency in two ways: by calculating the perplexity of \textit{GPT2-XL} and by using a CoLA grammaticality model to assess the grammaticality of the generated sentences. Diversity is evaluated in the same way as distinct scores. 

\paragraph{Results} The experimental results in table \ref{toxic_auto} show that CriticControl outperforms other existing baselines. CriticControl effectively avoids using toxic language while maintaining the natural flow of the text. Since this task aims to remove toxic elements while preserving the amount of information, the diversity scores are all similar between baselines.

\subsubsection{Human Evaluation} 

For human evaluation, we conduct preference tests comparing CriticControl to GPT2-large and DExperts. We randomly draw 200 samples from the results of the test set and ask annotators to indicate 1) which one is more rude or disrespectful (toxic comparison) and 2) which one is more grammatically correct and coherent? As in the topic and sentiment control experiment, we adopt the majority vote of three annotations for each comparison.

\paragraph{Results} Human evaluation results in \ref{fig:toxic_preference} state that Critic-Guided text generation is almost equivalent to the fluency of GPT2 and exceeds the DExperts. Furthermore, the results verify that when evaluating the toxicity of the text generated from the same prompts, CriticControl consistently outperforms both GPT2 and Dexperts.

\section{Conclusion}


We propose CriticControl, a controlled text generation method that takes advantage of both reinforcement learning and weighted decoding. 
Through experiments on various tasks and human studies, we demonstrate that CriticControl can effectively guide language models toward desired attributes while producing high-quality texts. Additionally, we show that CriticControl has a strong generalization ability in zero-shot attribute control tasks by using a general reward model. However, one of the limitations is its high computational cost to explore with `GPT3-scale' language models~\cite{brown2020language}, which may be addressed through offline reinforcement learning~\cite{fujimoto2019off} techniques in future research. Overall, CriticControl is promising for potential use in various forms of controlled text generation in future work.

\section*{Acknowledgements}
K. Jung is with ASRI, Seoul National University, Korea. This work has been financially supported by SNU-NAVER Hyperscale AI Center.

\bibliography{anthology,custom}

\begin{thebibliography}{51}
\expandafter\ifx\csname natexlab\endcsname\relax\def\natexlab#1{#1}\fi

\bibitem[{Arora et~al.(2022)Arora, Shuster, Sukhbaatar, and
  Weston}]{arora2022director}
Kushal Arora, Kurt Shuster, Sainbayar Sukhbaatar, and Jason Weston. 2022.
\newblock Director: Generator-classifiers for supervised language modeling.
\newblock \emph{arXiv preprint arXiv:2206.07694}.

\bibitem[{Bahdanau et~al.(2016)Bahdanau, Brakel, Xu, Goyal, Lowe, Pineau,
  Courville, and Bengio}]{bahdanau2016actor}
Dzmitry Bahdanau, Philemon Brakel, Kelvin Xu, Anirudh Goyal, Ryan Lowe, Joelle
  Pineau, Aaron Courville, and Yoshua Bengio. 2016.
\newblock An actor-critic algorithm for sequence prediction.
\newblock \emph{arXiv preprint arXiv:1607.07086}.

\bibitem[{Brown et~al.(2020)Brown, Mann, Ryder, Subbiah, Kaplan, Dhariwal,
  Neelakantan, Shyam, Sastry, Askell et~al.}]{brown2020language}
Tom Brown, Benjamin Mann, Nick Ryder, Melanie Subbiah, Jared~D Kaplan, Prafulla
  Dhariwal, Arvind Neelakantan, Pranav Shyam, Girish Sastry, Amanda Askell,
  et~al. 2020.
\newblock Language models are few-shot learners.
\newblock \emph{Advances in neural information processing systems},
  33:1877--1901.

\bibitem[{Dathathri et~al.(2019)Dathathri, Madotto, Lan, Hung, Frank, Molino,
  Yosinski, and Liu}]{dathathri2019plug}
Sumanth Dathathri, Andrea Madotto, Janice Lan, Jane Hung, Eric Frank, Piero
  Molino, Jason Yosinski, and Rosanne Liu. 2019.
\newblock Plug and play language models: A simple approach to controlled text
  generation.
\newblock \emph{arXiv preprint arXiv:1912.02164}.

\bibitem[{Devlin et~al.(2018)Devlin, Chang, Lee, and
  Toutanova}]{devlin2018bert}
Jacob Devlin, Ming-Wei Chang, Kenton Lee, and Kristina Toutanova. 2018.
\newblock Bert: Pre-training of deep bidirectional transformers for language
  understanding.
\newblock \emph{arXiv preprint arXiv:1810.04805}.

\bibitem[{Fan et~al.(2018)Fan, Lewis, and Dauphin}]{fan2018hierarchical}
Angela Fan, Mike Lewis, and Yann Dauphin. 2018.
\newblock Hierarchical neural story generation.
\newblock \emph{arXiv preprint arXiv:1805.04833}.

\bibitem[{Fujimoto et~al.(2019)Fujimoto, Meger, and Precup}]{fujimoto2019off}
Scott Fujimoto, David Meger, and Doina Precup. 2019.
\newblock Off-policy deep reinforcement learning without exploration.
\newblock In \emph{International conference on machine learning}, pages
  2052--2062. PMLR.

\bibitem[{Gehman et~al.(2020)Gehman, Gururangan, Sap, Choi, and
  Smith}]{gehman2020realtoxicityprompts}
Samuel Gehman, Suchin Gururangan, Maarten Sap, Yejin Choi, and Noah~A Smith.
  2020.
\newblock Realtoxicityprompts: Evaluating neural toxic degeneration in language
  models.
\newblock \emph{arXiv preprint arXiv:2009.11462}.

\bibitem[{Ghazvininejad et~al.(2017)Ghazvininejad, Shi, Priyadarshi, and
  Knight}]{ghazvininejad2017hafez}
Marjan Ghazvininejad, Xing Shi, Jay Priyadarshi, and Kevin Knight. 2017.
\newblock Hafez: an interactive poetry generation system.
\newblock In \emph{Proceedings of ACL 2017, System Demonstrations}, pages
  43--48.

\bibitem[{Gong et~al.(2019)Gong, Bhat, Wu, Xiong, and
  Hwu}]{gong2019reinforcement}
Hongyu Gong, Suma Bhat, Lingfei Wu, JinJun Xiong, and Wen-mei Hwu. 2019.
\newblock Reinforcement learning based text style transfer without parallel
  training corpus.
\newblock \emph{arXiv preprint arXiv:1903.10671}.

\bibitem[{Greensmith et~al.(2004)Greensmith, Bartlett, and
  Baxter}]{greensmith2004variance}
Evan Greensmith, Peter~L Bartlett, and Jonathan Baxter. 2004.
\newblock Variance reduction techniques for gradient estimates in reinforcement
  learning.
\newblock \emph{Journal of Machine Learning Research}, 5(9).

\bibitem[{Gu et~al.(2022)Gu, Feng, Ma, Zhang, Gong, and
  Qin}]{gu2022distributional}
Yuxuan Gu, Xiaocheng Feng, Sicheng Ma, Lingyuan Zhang, Heng Gong, and Bing Qin.
  2022.
\newblock A distributional lens for multi-aspect controllable text generation.
\newblock \emph{arXiv preprint arXiv:2210.02889}.

\bibitem[{Guo et~al.(2021)Guo, Tan, Liu, Xing, and Hu}]{guo2021text}
Han Guo, Bowen Tan, Zhengzhong Liu, Eric~P Xing, and Zhiting Hu. 2021.
\newblock Text generation with efficient (soft) q-learning.
\newblock \emph{arXiv preprint arXiv:2106.07704}.

\bibitem[{Gururangan et~al.(2020)Gururangan, Marasovi{\'c}, Swayamdipta, Lo,
  Beltagy, Downey, and Smith}]{gururangan2020don}
Suchin Gururangan, Ana Marasovi{\'c}, Swabha Swayamdipta, Kyle Lo, Iz~Beltagy,
  Doug Downey, and Noah~A Smith. 2020.
\newblock Don't stop pretraining: adapt language models to domains and tasks.
\newblock \emph{arXiv preprint arXiv:2004.10964}.

\bibitem[{Holtzman et~al.(2019)Holtzman, Buys, Du, Forbes, and
  Choi}]{holtzman2019curious}
Ari Holtzman, Jan Buys, Li~Du, Maxwell Forbes, and Yejin Choi. 2019.
\newblock The curious case of neural text degeneration.
\newblock \emph{arXiv preprint arXiv:1904.09751}.

\bibitem[{Holtzman et~al.(2018)Holtzman, Buys, Forbes, Bosselut, Golub, and
  Choi}]{holtzman2018learning}
Ari Holtzman, Jan Buys, Maxwell Forbes, Antoine Bosselut, David Golub, and
  Yejin Choi. 2018.
\newblock Learning to write with cooperative discriminators.
\newblock \emph{arXiv preprint arXiv:1805.06087}.

\bibitem[{Hosseini et~al.(2017)Hosseini, Kannan, Zhang, and
  Poovendran}]{hosseini2017deceiving}
Hossein Hosseini, Sreeram Kannan, Baosen Zhang, and Radha Poovendran. 2017.
\newblock Deceiving google's perspective api built for detecting toxic
  comments.
\newblock \emph{arXiv preprint arXiv:1702.08138}.

\bibitem[{Hu et~al.(2017)Hu, Yang, Liang, Salakhutdinov, and
  Xing}]{hu2017toward}
Zhiting Hu, Zichao Yang, Xiaodan Liang, Ruslan Salakhutdinov, and Eric~P Xing.
  2017.
\newblock Toward controlled generation of text.
\newblock In \emph{International conference on machine learning}, pages
  1587--1596. PMLR.

\bibitem[{Jang et~al.(2021)Jang, Lee, and Kim}]{jang2021gpt}
Youngsoo Jang, Jongmin Lee, and Kee-Eung Kim. 2021.
\newblock Gpt-critic: Offline reinforcement learning for end-to-end
  task-oriented dialogue systems.
\newblock In \emph{International Conference on Learning Representations}.

\bibitem[{Keskar et~al.(2019)Keskar, McCann, Varshney, Xiong, and
  Socher}]{keskar2019ctrl}
Nitish~Shirish Keskar, Bryan McCann, Lav~R Varshney, Caiming Xiong, and Richard
  Socher. 2019.
\newblock Ctrl: A conditional transformer language model for controllable
  generation.
\newblock \emph{arXiv preprint arXiv:1909.05858}.

\bibitem[{Kim et~al.(2021)Kim, Kim, Lee, Lee, Kwak, Jeon, Park, Kim, Kim, Seo
  et~al.}]{kim2021changes}
Boseop Kim, HyoungSeok Kim, Sang-Woo Lee, Gichang Lee, Donghyun Kwak,
  Dong~Hyeon Jeon, Sunghyun Park, Sungju Kim, Seonhoon Kim, Dongpil Seo, et~al.
  2021.
\newblock What changes can large-scale language models bring? intensive study
  on hyperclova: Billions-scale korean generative pretrained transformers.
\newblock \emph{arXiv preprint arXiv:2109.04650}.

\bibitem[{Krause et~al.(2020)Krause, Gotmare, McCann, Keskar, Joty, Socher, and
  Rajani}]{krause2020gedi}
Ben Krause, Akhilesh~Deepak Gotmare, Bryan McCann, Nitish~Shirish Keskar,
  Shafiq Joty, Richard Socher, and Nazneen~Fatema Rajani. 2020.
\newblock Gedi: Generative discriminator guided sequence generation.
\newblock \emph{arXiv preprint arXiv:2009.06367}.

\bibitem[{Kumar et~al.(2021)Kumar, Malmi, Severyn, and
  Tsvetkov}]{kumar2021controlled}
Sachin Kumar, Eric Malmi, Aliaksei Severyn, and Yulia Tsvetkov. 2021.
\newblock Controlled text generation as continuous optimization with multiple
  constraints.
\newblock \emph{Advances in Neural Information Processing Systems},
  34:14542--14554.

\bibitem[{Lewis et~al.(2019)Lewis, Liu, Goyal, Ghazvininejad, Mohamed, Levy,
  Stoyanov, and Zettlemoyer}]{lewis2019bart}
Mike Lewis, Yinhan Liu, Naman Goyal, Marjan Ghazvininejad, Abdelrahman Mohamed,
  Omer Levy, Ves Stoyanov, and Luke Zettlemoyer. 2019.
\newblock Bart: Denoising sequence-to-sequence pre-training for natural
  language generation, translation, and comprehension.
\newblock \emph{arXiv preprint arXiv:1910.13461}.

\bibitem[{Li et~al.(2015)Li, Galley, Brockett, Gao, and
  Dolan}]{li2015diversity}
Jiwei Li, Michel Galley, Chris Brockett, Jianfeng Gao, and Bill Dolan. 2015.
\newblock A diversity-promoting objective function for neural conversation
  models.
\newblock \emph{arXiv preprint arXiv:1510.03055}.

\bibitem[{Li et~al.(2016)Li, Monroe, Ritter, Galley, Gao, and
  Jurafsky}]{li2016deep}
Jiwei Li, Will Monroe, Alan Ritter, Michel Galley, Jianfeng Gao, and Dan
  Jurafsky. 2016.
\newblock Deep reinforcement learning for dialogue generation.
\newblock \emph{arXiv preprint arXiv:1606.01541}.

\bibitem[{Liu et~al.(2021)Liu, Sap, Lu, Swayamdipta, Bhagavatula, Smith, and
  Choi}]{liu2021dexperts}
Alisa Liu, Maarten Sap, Ximing Lu, Swabha Swayamdipta, Chandra Bhagavatula,
  Noah~A Smith, and Yejin Choi. 2021.
\newblock Dexperts: Decoding-time controlled text generation with experts and
  anti-experts.
\newblock \emph{arXiv preprint arXiv:2105.03023}.

\bibitem[{Lu et~al.(2022)Lu, Welleck, Jiang, Hessel, Qin, West, Ammanabrolu,
  and Choi}]{lu2022quark}
Ximing Lu, Sean Welleck, Liwei Jiang, Jack Hessel, Lianhui Qin, Peter West,
  Prithviraj Ammanabrolu, and Yejin Choi. 2022.
\newblock Quark: Controllable text generation with reinforced unlearning.
\newblock \emph{arXiv preprint arXiv:2205.13636}.

\bibitem[{Mireshghallah et~al.(2022)Mireshghallah, Goyal, and
  Berg-Kirkpatrick}]{mireshghallah2022mix}
Fatemehsadat Mireshghallah, Kartik Goyal, and Taylor Berg-Kirkpatrick. 2022.
\newblock Mix and match: Learning-free controllable text generation using
  energy language models.
\newblock \emph{arXiv preprint arXiv:2203.13299}.

\bibitem[{Nakano et~al.(2021)Nakano, Hilton, Balaji, Wu, Ouyang, Kim, Hesse,
  Jain, Kosaraju, Saunders et~al.}]{nakano2021webgpt}
Reiichiro Nakano, Jacob Hilton, Suchir Balaji, Jeff Wu, Long Ouyang, Christina
  Kim, Christopher Hesse, Shantanu Jain, Vineet Kosaraju, William Saunders,
  et~al. 2021.
\newblock Webgpt: Browser-assisted question-answering with human feedback.
\newblock \emph{arXiv preprint arXiv:2112.09332}.

\bibitem[{Nguyen et~al.(2017)Nguyen, Daum{\'e}~III, and
  Boyd-Graber}]{nguyen2017reinforcement}
Khanh Nguyen, Hal Daum{\'e}~III, and Jordan Boyd-Graber. 2017.
\newblock Reinforcement learning for bandit neural machine translation with
  simulated human feedback.
\newblock \emph{arXiv preprint arXiv:1707.07402}.

\bibitem[{Paulus et~al.(2017)Paulus, Xiong, and Socher}]{paulus2017deep}
Romain Paulus, Caiming Xiong, and Richard Socher. 2017.
\newblock A deep reinforced model for abstractive summarization.
\newblock \emph{arXiv preprint arXiv:1705.04304}.

\bibitem[{Radford et~al.(2019)Radford, Wu, Child, Luan, Amodei, Sutskever
  et~al.}]{radford2019language}
Alec Radford, Jeffrey Wu, Rewon Child, David Luan, Dario Amodei, Ilya
  Sutskever, et~al. 2019.
\newblock Language models are unsupervised multitask learners.
\newblock \emph{OpenAI blog}, 1(8):9.

\bibitem[{Ranzato et~al.(2015)Ranzato, Chopra, Auli, and
  Zaremba}]{ranzato2015sequence}
Marc'Aurelio Ranzato, Sumit Chopra, Michael Auli, and Wojciech Zaremba. 2015.
\newblock Sequence level training with recurrent neural networks.
\newblock \emph{arXiv preprint arXiv:1511.06732}.

\bibitem[{Sanh et~al.(2019)Sanh, Debut, Chaumond, and
  Wolf}]{sanh2019distilbert}
Victor Sanh, Lysandre Debut, Julien Chaumond, and Thomas Wolf. 2019.
\newblock Distilbert, a distilled version of bert: smaller, faster, cheaper and
  lighter.
\newblock \emph{arXiv preprint arXiv:1910.01108}.

\bibitem[{Schulman et~al.(2015)Schulman, Moritz, Levine, Jordan, and
  Abbeel}]{schulman2015high}
John Schulman, Philipp Moritz, Sergey Levine, Michael Jordan, and Pieter
  Abbeel. 2015.
\newblock High-dimensional continuous control using generalized advantage
  estimation.
\newblock \emph{arXiv preprint arXiv:1506.02438}.

\bibitem[{Schulman et~al.(2017)Schulman, Wolski, Dhariwal, Radford, and
  Klimov}]{schulman2017proximal}
John Schulman, Filip Wolski, Prafulla Dhariwal, Alec Radford, and Oleg Klimov.
  2017.
\newblock Proximal policy optimization algorithms.
\newblock \emph{arXiv preprint arXiv:1707.06347}.

\bibitem[{Sharma et~al.(2021)Sharma, Lin, Miner, Atkins, and
  Althoff}]{sharma2021towards}
Ashish Sharma, Inna~W Lin, Adam~S Miner, David~C Atkins, and Tim Althoff. 2021.
\newblock Towards facilitating empathic conversations in online mental health
  support: A reinforcement learning approach.
\newblock In \emph{Proceedings of the Web Conference 2021}, pages 194--205.

\bibitem[{Snell et~al.(2022)Snell, Kostrikov, Su, Yang, and
  Levine}]{snell2022offline}
Charlie Snell, Ilya Kostrikov, Yi~Su, Mengjiao Yang, and Sergey Levine. 2022.
\newblock Offline rl for natural language generation with implicit language q
  learning.
\newblock \emph{arXiv preprint arXiv:2206.11871}.

\bibitem[{Stiennon et~al.(2020)Stiennon, Ouyang, Wu, Ziegler, Lowe, Voss,
  Radford, Amodei, and Christiano}]{stiennon2020learning}
Nisan Stiennon, Long Ouyang, Jeffrey Wu, Daniel Ziegler, Ryan Lowe, Chelsea
  Voss, Alec Radford, Dario Amodei, and Paul~F Christiano. 2020.
\newblock Learning to summarize with human feedback.
\newblock \emph{Advances in Neural Information Processing Systems},
  33:3008--3021.

\bibitem[{Sudhakar et~al.(2019)Sudhakar, Upadhyay, and
  Maheswaran}]{sudhakar-etal-2019-transforming}
Akhilesh Sudhakar, Bhargav Upadhyay, and Arjun Maheswaran. 2019.
\newblock \href {https://doi.org/10.18653/v1/D19-1322} {{``}transforming{''}
  delete, retrieve, generate approach for controlled text style transfer}.
\newblock In \emph{Proceedings of the 2019 Conference on Empirical Methods in
  Natural Language Processing and the 9th International Joint Conference on
  Natural Language Processing (EMNLP-IJCNLP)}, pages 3269--3279, Hong Kong,
  China. Association for Computational Linguistics.

\bibitem[{Sutton and Barto(2018)}]{sutton2018reinforcement}
Richard~S Sutton and Andrew~G Barto. 2018.
\newblock \emph{Reinforcement learning: An introduction}.
\newblock MIT press.

\bibitem[{Sutton et~al.(1999)Sutton, McAllester, Singh, and
  Mansour}]{sutton1999policy}
Richard~S Sutton, David McAllester, Satinder Singh, and Yishay Mansour. 1999.
\newblock Policy gradient methods for reinforcement learning with function
  approximation.
\newblock \emph{Advances in neural information processing systems}, 12.

\bibitem[{Upadhyay et~al.(2022)Upadhyay, Sudhakar, and
  Maheswaran}]{upadhyay2022efficient}
Bhargav Upadhyay, Akhilesh Sudhakar, and Arjun Maheswaran. 2022.
\newblock Efficient reinforcement learning for unsupervised controlled text
  generation.
\newblock \emph{arXiv preprint arXiv:2204.07696}.

\bibitem[{Warstadt et~al.(2019)Warstadt, Singh, and
  Bowman}]{warstadt2019neural}
Alex Warstadt, Amanpreet Singh, and Samuel~R Bowman. 2019.
\newblock Neural network acceptability judgments.
\newblock \emph{Transactions of the Association for Computational Linguistics},
  7:625--641.

\bibitem[{Wu et~al.(2021)Wu, Ouyang, Ziegler, Stiennon, Lowe, Leike, and
  Christiano}]{wu2021recursively}
Jeff Wu, Long Ouyang, Daniel~M Ziegler, Nisan Stiennon, Ryan Lowe, Jan Leike,
  and Paul Christiano. 2021.
\newblock Recursively summarizing books with human feedback.
\newblock \emph{arXiv preprint arXiv:2109.10862}.

\bibitem[{Wu et~al.(2016)Wu, Schuster, Chen, Le, Norouzi, Macherey, Krikun,
  Cao, Gao, Macherey et~al.}]{wu2016google}
Yonghui Wu, Mike Schuster, Zhifeng Chen, Quoc~V Le, Mohammad Norouzi, Wolfgang
  Macherey, Maxim Krikun, Yuan Cao, Qin Gao, Klaus Macherey, et~al. 2016.
\newblock Google's neural machine translation system: Bridging the gap between
  human and machine translation.
\newblock \emph{arXiv preprint arXiv:1609.08144}.

\bibitem[{Wu and Hu(2018)}]{wu2018learning}
Yuxiang Wu and Baotian Hu. 2018.
\newblock Learning to extract coherent summary via deep reinforcement learning.
\newblock In \emph{Proceedings of the AAAI Conference on Artificial
  Intelligence}, volume~32.

\bibitem[{Yang and Klein(2021)}]{yang-klein-2021-fudge}
Kevin Yang and Dan Klein. 2021.
\newblock \href {https://doi.org/10.18653/v1/2021.naacl-main.276} {{FUDGE}:
  Controlled text generation with future discriminators}.
\newblock In \emph{Proceedings of the 2021 Conference of the North American
  Chapter of the Association for Computational Linguistics: Human Language
  Technologies}, pages 3511--3535, Online. Association for Computational
  Linguistics.

\bibitem[{Yi et~al.(2019)Yi, Goel, Khatri, Cervone, Chung, Hedayatnia,
  Venkatesh, Gabriel, and Hakkani-Tur}]{yi2019towards}
Sanghyun Yi, Rahul Goel, Chandra Khatri, Alessandra Cervone, Tagyoung Chung,
  Behnam Hedayatnia, Anu Venkatesh, Raefer Gabriel, and Dilek Hakkani-Tur.
  2019.
\newblock Towards coherent and engaging spoken dialog response generation using
  automatic conversation evaluators.
\newblock \emph{arXiv preprint arXiv:1904.13015}.

\bibitem[{Ziegler et~al.(2019)Ziegler, Stiennon, Wu, Brown, Radford, Amodei,
  Christiano, and Irving}]{ziegler2019fine}
Daniel~M Ziegler, Nisan Stiennon, Jeffrey Wu, Tom~B Brown, Alec Radford, Dario
  Amodei, Paul Christiano, and Geoffrey Irving. 2019.
\newblock Fine-tuning language models from human preferences.
\newblock \emph{arXiv preprint arXiv:1909.08593}.

\end{thebibliography}
\bibliographystyle{acl_natbib}

\appendix

\end{document}